\pdfoutput=1

\documentclass[11pt]{article}

\usepackage{acl}

\usepackage{times}
\usepackage{latexsym}

\usepackage[T1]{fontenc}

\usepackage[utf8]{inputenc}

\usepackage{microtype}

\usepackage{inconsolata}

\usepackage{refstyle}
\usepackage{amsmath}
\usepackage{cleveref}
\usepackage{colortbl}

\usepackage{booktabs}
\usepackage{multirow} %
\usepackage{soul}%
\usepackage{tabularx}

\usepackage{pifont}

\usepackage{graphicx}

\crefformat{section}{\S#2#1#3}
\crefformat{subsection}{\S#2#1#3}
\crefformat{subsubsection}{\S#2#1#3}
\crefrangeformat{section}{\S#3#1#4 to~\S#5#2#6}
\crefmultiformat{section}{\S#2#1#3}{ and~\S#2#1#3}{, #2#1#3}{ and~#2#1#3}
\Crefformat{figure}{#2Fig.~#1#3}
\Crefmultiformat{figure}{Figs.~#2#1#3}{ and~#2#1#3}{, #2#1#3}{ and~#2#1#3}
\Crefformat{table}{#2Tab.~#1#3}
\Crefmultiformat{table}{Tabs.~#2#1#3}{ and~#2#1#3}{, #2#1#3}{ and~#2#1#3}
\Crefformat{appendix}{#2Appx.~\S#1#3}
\crefformat{algorithm}{Alg.~#2#1#3}
\Crefformat{equation}{#2Eq.~#1#3}

\newcommand{\stitle}[1]{\vspace{1ex} \noindent{\bf #1.}}

\usepackage{xspace}
\newcommand{\ie}{\textit{i}.\textit{e}.}

\newcommand{\piref}{\pi_\text{ref}}

\newcommand{\MODEL}{\mbox{\textsc{mDPO}}\xspace}

\title{
\MODEL: Conditional Preference Optimization for \\ 
Multimodal Large Language Models
}

\author{
Fei Wang$^1$~~
Wenxuan Zhou$^1$~~
James Y. Huang$^1$~~
Nan Xu$^1$~~\\
\textbf{Sheng Zhang}$^2$~~
\textbf{Hoifung Poon}$^2$~~
\textbf{Muhao Chen}$^3$ \\
[5pt]
$^1$University of Southern California~~
$^2$Microsoft Research~~
$^3$University of California, Davis \\
[5pt]
\url{https://feiwang96.github.io/mDPO} \\
[5pt]
\texttt{
\{fwang598,zhouwenx\}@usc.edu~~
shezhan@microsoft.com~~
muhchen@ucdavis.edu
} 
}

\begin{document}
\maketitle

\begin{abstract}
    Direct preference optimization (DPO) has shown to be an effective method for large language model (LLM) alignment. Recent works have attempted to apply DPO to multimodal scenarios but have found it challenging to achieve consistent improvement. Through a comparative experiment, we identify the \textit{unconditional preference} problem in multimodal preference optimization, where the model overlooks the image condition. To address this problem, we propose \MODEL, a multimodal DPO objective that prevents the over-prioritization of language-only preferences by also optimizing image preference. Moreover, we introduce a reward anchor that forces the reward to be positive for chosen responses, thereby avoiding the decrease in their likelihood—an intrinsic problem of \textit{relative preference} optimization. Experiments on two multimodal LLMs of different sizes and three widely used benchmarks demonstrate that \MODEL effectively addresses the unconditional preference problem in multimodal preference optimization and significantly improves model performance, particularly in reducing hallucination.%
\end{abstract}

\section{Introduction}

Direct preference optimization (DPO) has emerged as the predominating method for aligning large language models (LLMs) with human preferences~\cite{rafailov2024direct,zhao2024survey}. Building on its success in the language modality, recent studies have extended DPO to multimodal scenarios~\cite{li2023silkie,yu2024rlhf,zhou2024aligning,zhao2023beyond}. However, transferring this approach across modalities presents significant challenges. Merely substituting textual preference data with multimodal preference data does not consistently yield positive outcomes and can exacerbate issues such as hallucinations \cite{li2023silkie,sarkar2024mitigating}.

\begin{figure}[t]
    \centering
    \includegraphics[width=0.9\linewidth]{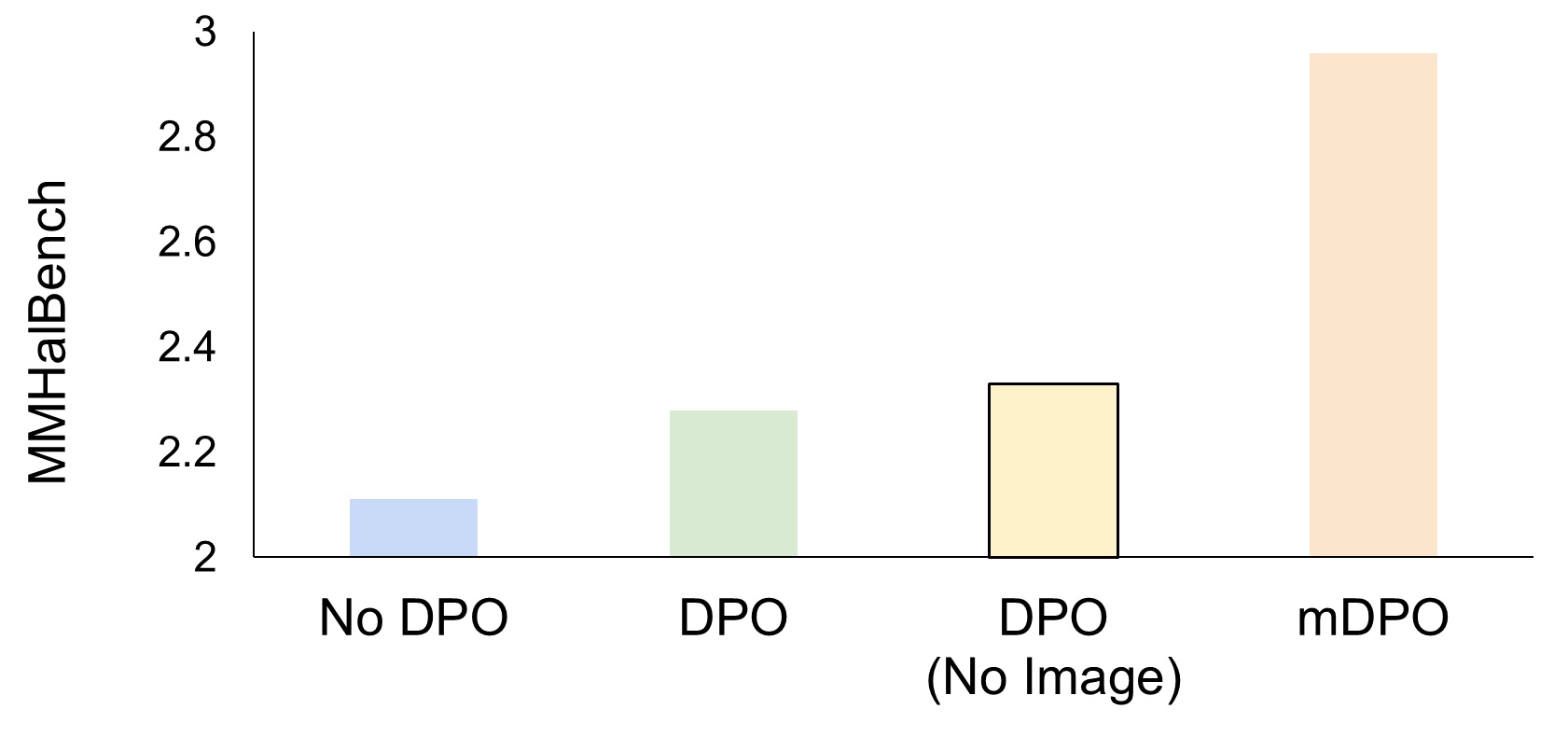}
    \caption{We train Bunny-v1.0-3B \cite{he2024bunny} on 10K multimodal preference data from Silkie \cite{li2023silkie} with different variants of DPO. We perform DPO (No Image) where all images are removed from the preference data. Counterintuitively, the overall score on the MMHalBench \cite{sun2023aligning} for DPO (No Image) is similar to that of DPO with images. This finding suggests that DPO may suffer from unconditional preferences, neglecting the visual modality during optimization. Our proposed method, \MODEL, effectively addresses this issue and improves model performance.}
    \label{fig:problem}
\end{figure}

\begin{figure*}[t]
    \centering
    \includegraphics[width=1\linewidth]{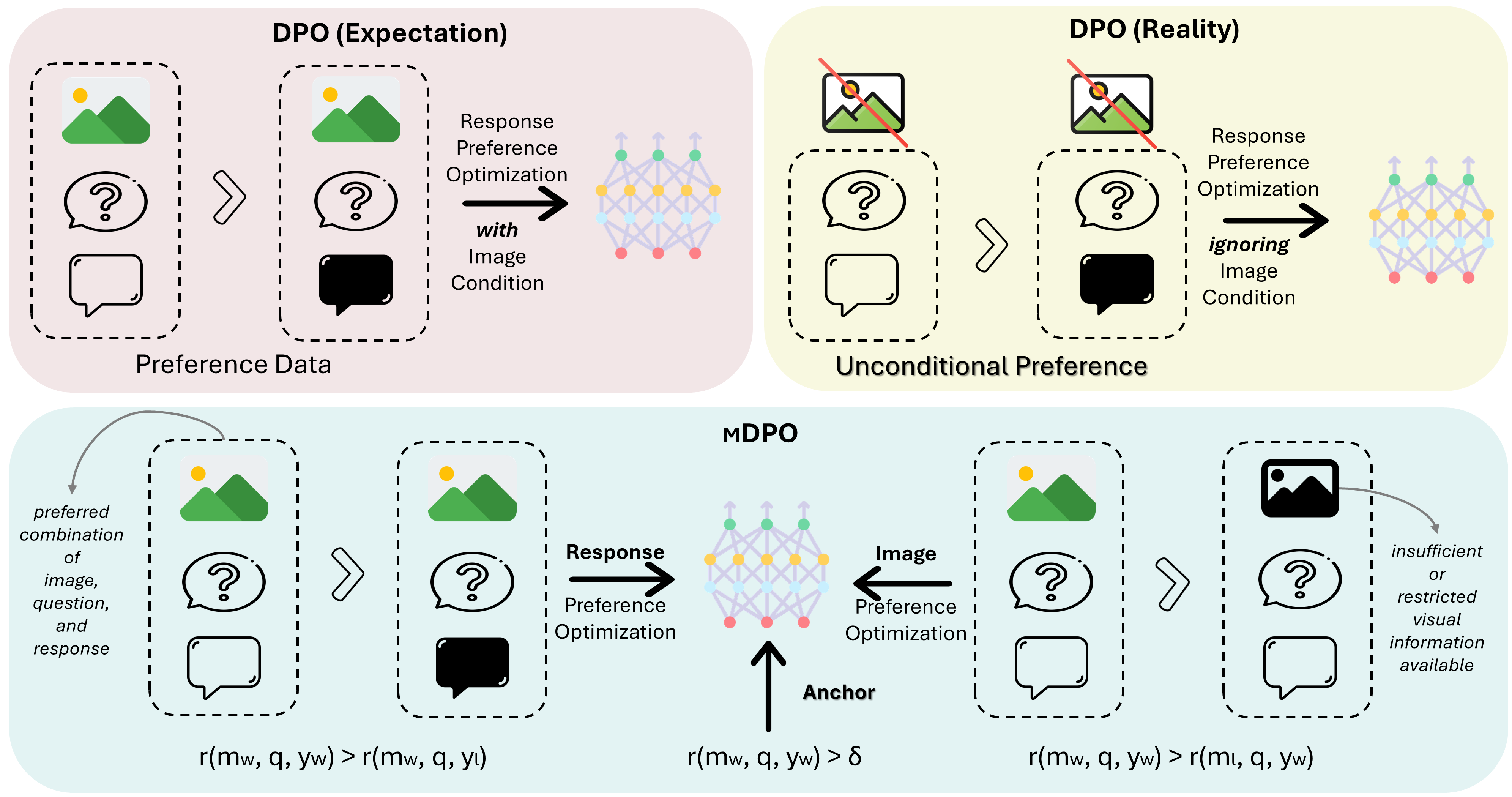}
    \caption{\textbf{Overview of \MODEL}. \textit{Top Left}: Standard DPO expects the multimodal LLM to learn response preferences conditioned on both the image and the question. \textit{Top Right}: However, in practice, the learning process often disregards the image condition. \textit{Bottom}: To address this issue, \MODEL introduces an additional image preference learning objective to emphasize the relationship between the image and the response. Furthermore, \MODEL incorporates a reward anchor to ensure that the probability of the chosen response does not decrease.}
    \label{fig:mdpo}
\end{figure*}

\begin{figure*}[t]
    \centering
    \includegraphics[width=1\linewidth]{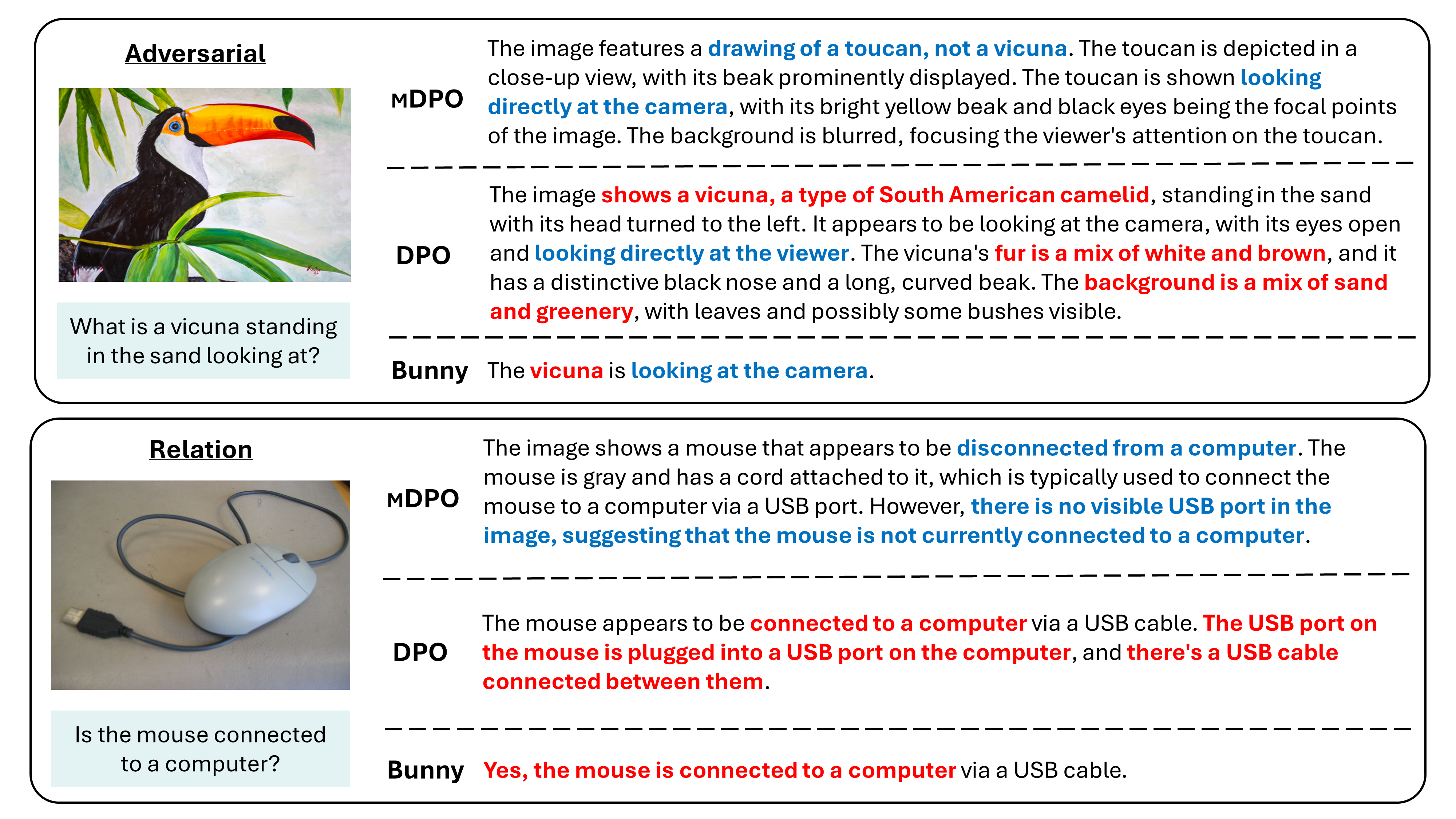}
    \caption{\textbf{Qualitative Results} from MMHalBench. \textit{Top}: When trained with standard DPO, Bunny often assumes the image description in the question is correct, responding accordingly, even if the question contains an adversarial premise regarding the image. In contrast, \MODEL identifies the false premise in the question by referencing the image. \textit{Bottom}: Bunny trained with standard DPO may disregard the image and provide an educated guess for the answer. Conversely, \MODEL delivers a correct answer that is conditioned on the image.}
    \label{fig:qual}
\end{figure*}

While recent efforts in multimodal preference learning focus on improving performance through enhanced multimodal preference data \cite{li2023silkie,zhao2023beyond,xiao2024detecting,zhou2024aligning,pi2024strengthening,sarkar2024mitigating,yu2024rlaif,deng2024enhancing}, we investigate the pitfalls of multimodal DPO from a different perspective. Through controlled comparisons, we discover that multimodal LLMs can achieve similar performance even when all images are \emph{removed} from the multimodal preference data during DPO (see \Cref{fig:problem}). This counterintuitive finding suggests that the failure of DPO in multimodal scenarios may not be solely attributed to data quality. We attribute this to a systematic gap between the theoretical expectations and practical implementations of the DPO objective in multimodal settings (refer to \Cref{fig:mdpo}). While DPO aims to compute implicit rewards conditioned on all input modalities, it may prioritize language-only preferences and overlook the image condition (\ie, \emph{unconditional preference}),  leading to suboptimal model performance and increased hallucination. After applying DPO, the model may show an increased tendency to ignore the provided image and generate responses based solely on the question (illustrated in \Cref{fig:qual}).

In this paper, we propose \MODEL, a multimodal DPO objective that utilizes conditional preference optimization on images to prevent the overly prioritization of language-only preferences. As depicted in \Cref{fig:mdpo}, in addition to the original preference pairs contrasting responses, \MODEL introduces new preference pairs contrasting images. The rejected image is derived from the original (\ie, chosen) image by reducing effective visual information. This approach, combined with the standard DPO objective, encourages the multimodal LLM to simultaneously emphasize both visual and language features.
Furthermore, we observe that DPO often experiences a decrease in the likelihood of chosen responses in multimodal scenarios despite an increase in the implicit reward. To address this, \MODEL incorporates a reward anchor that maintains the likelihood of chosen responses by regularizing the reward to be positive.

To validate the effectiveness of \MODEL, we conduct experiments using Bunny-v1.0-3B \cite{he2024bunny} and LLaVA-v1.5-7B \cite{liu2024improved}. Both automatic and human evaluations on MMHalbench \cite{sun2023aligning}, Object HalBench \cite{yu2024rlhf}, and AMBER \cite{wang2023llm} consistently demonstrate that \MODEL outperforms standard DPO in multimodal scenarios, effectively reducing hallucinations across varying model and data scales. Detailed analyses reveal that conditional preference plays a crucial role in enhancing the effectiveness of DPO for multimodal LLMs. Fine-grained and qualitative studies further illustrate that \MODEL significantly improves the model's ability to comprehend images and mitigates language biases in model responses.

Our contributions are three-fold. First, we identify unconditional preference towards the visual modality as a primary reason for the pitfalls of DPO in multimodal LLMs. Second, we propose \MODEL, a multimodal DPO objective that incorporates conditional preference optimization and anchored preference optimization to mitigate these pitfalls. Third, we verify the effectiveness of \MODEL across different model and data scales. Using \MODEL, we achieve the best-performing 3B multimodal LLM in terms of reducing hallucinations.

\section{The Pitfall of Preference Optimization}

In this section, we first introduce the background of DPO (\Cref{sec:dpo}) and then delve into the issue of unconditional preference in multimodal DPO (\Cref{sec:unconditional}).

\subsection{Background: Preference Optimization}
\label{sec:dpo}

Preference optimization seeks to align LLMs with human preferences, thereby enhancing their capabilities to respond to human needs. In the context of LLMs, it aims to encourage the model to learn that, for a given question \( q \), the response \( y_w \) chosen by the evaluator is preferred over the rejected one \( y_l \). DPO is the predominant method for this purpose. Derived from reward modeling in RLHF~\cite{ouyang2022training}, it seeks to maximize the difference between the reward for the chosen response \( r(q, y_w) \) and that for the rejected response \( r(q, y_l) \).
Specifically, given a model to be optimized $\pi_\theta$ and a reference model $\pi_\text{ref}$, which is typically initialized from a supervised finetuning model, DPO formulates the reward as follows:

\begin{small}
$$\small r(q, y) = \beta \log \frac{\pi_{\theta}(y|q)}{\piref(y|q)} + Z(q),$$
\end{small}
where $Z(q)$ is a partition function, $\beta$ is a hyperparameter that controls the deviation from the reference model.
Then, based on the Bradley-Terry model~\cite{bradley1952rank}, the preference optimization objective becomes:
\begin{equation*}
\resizebox{\linewidth}{!}{
    $\mathcal{L}_\text{DPO} = -\log \sigma \left(\beta \log \frac{\pi_{\theta}(y_w | q)}{\piref(y_w | q)} - \beta \log \frac{\pi_{\theta}(y_l | q)}{\piref(y_l | q)}\right),$}
\end{equation*}
which is essentially maximizing 

\begin{small}
$$\sigma(r(q, y_w) - r(q, y_l)).$$
\end{small}

In the multimodal scenario, each instance in the preference data contains an image $m$, in addition to $q$, $y_w$, and  $y_l$, and the preference label is decided based on both the image and the question. DPO expects the multimodal LLM to learn to maximize

\begin{small}
$$\sigma(r(m, q, y_w) - r(m, q, y_l)),$$
\end{small}
and the objective becomes:
\begin{equation*}
\resizebox{\linewidth}{!}{
         $\mathcal{L}_{\text{DPO}_m} = -\log \sigma \Big(\beta \log \frac{\pi_{\theta}(y_w| m, q)}{\piref(y_w| m, q)} - \beta \log \frac{\pi_{\theta}(y_l| m, q)}{\piref(y_l| m, q)}\Big).$
}
\end{equation*}

\subsection{Problem: Unconditional Preference}
\label{sec:unconditional}

Recent studies have found inconsistent improvements in model capabilities when applying DPO to multimodal LLMs, often attributing this issue to the quality of preference data \cite{li2023silkie, sarkar2024mitigating}. However, our controlled experiments suggest that the problem arises because DPO does not effectively utilize the visual modality in the preference dataset.
To explore this, we introduce a variant called DPO (No Image), which is trained on the preference dataset with the visual signal removed, forcing the model to maximize $\sigma(r(q, y_w) - r(q, y_l))$ without visual cues. We apply both DPO and DPO (No Image) to the Bunny-v1.0-3B model and 10K preference instances from the LLaVA-Instruct \cite{liu2024visual} subset of Silkie \cite{li2023silkie}.
Results shown in~\Cref{fig:problem} indicate that DPO (No Image) performs similarly, or even slightly better, than DPO on MMHalBench. This finding underscores the issue of \textit{unconditional preference} learned by multimodal LLMs during DPO, where the model may disregard image information, as illustrated in~\Cref{fig:mdpo}.

\section{\MODEL}
In this section, we introduce \MODEL, an improved DPO approached dedicated to multimodal preference alignment. As depicted in \Cref{fig:mdpo}, \MODEL introduces two additional preference optimization objectives to DPO: \emph{conditional preference optimization} to address the issue of ignoring visual information (see \Cref{sec:conditional}), and \emph{anchored preference optimization} to prevent a decrease in the likelihood of the chosen response (see \Cref{sec:anchored}).

\subsection{Conditional Preference Optimization}
\label{sec:conditional}

We propose a conditional preference optimization objective to address the issue of ignoring visual information in preference data. The core idea is to construct preference data where the image is the only variable, forcing the model to determine the preference label based on visual information.
Specifically, given a pair of tuples $(m_w, q, y_w)$ and $(m_l, q, y_w)$, where $m_w$ is more compitible with $q$ and $y_w$ than $m_l$, the conditional preference optimization objective is formulated as:
\begin{equation*}
\resizebox{\linewidth}{!}{$
\mathcal{L}_\text{CoPO} = -\log \sigma \Big(\beta \log \frac{\pi_{\theta}(y_w | m_w, q)}{\piref(y_w | m_w, q)} - \beta \log \frac{\pi_{\theta}(y_w | m_l, q)}{\piref(y_w | m_l, q)}\Big).$}
\end{equation*}
The challenge then lies in constructing appropriate pairs of $m_w$ and $m_l$. On the one hand, $m_l$ should contain different visual information from $m_w$ to make it less compatible. On the other hand, it should also share some common features with $m_w$ to serve as a hard negative.
We find that a straightforward strategy, using the original image as $m$ and creating $m_l$ by randomly cropping less than 20\% of the original image, yields the best performance, as shown in~\Cref{sec:analysis}.

To summarize, the standard DPO objective maximizes \( \sigma(r(m_w, q, y_w) - r(m_w, q, y_l)) \), while the conditional preference optimization objective maximizes \( \sigma(r(m_w, q, y_w) - r(m_l, q, y_w)) \). The two objectives work in collaboration to ensure that the multimodal LLM captures preferences based on both visual and language cues.
Notably, while we focus on the multimodal setting, this conditional preference optimization could be beneficial to other preference optimization scenarios involving multiple input components. In such settings, DPO may also ignore specific input components and encourage the model to learn unconditional preferences.

\begin{table*}[t]
\centering
\small
\setlength{\tabcolsep}{2.5pt}
\begin{tabular}{lcccccccc}\toprule
 &\multicolumn{2}{c}{MMHalBench} &\multicolumn{2}{c}{Object HalBench} &\multicolumn{4}{c}{AMBER} \\ \cmidrule(lr){2-3} \cmidrule(lr){4-5} \cmidrule(lr){6-9}
 &Score $\uparrow$ &HalRate $\downarrow$ &CHAIR$_{s}$ $\downarrow$&CHAIR$_{i}$ $\downarrow$&CHAIR$_{s}$ $\downarrow$&Cover. $\uparrow$ &HalRate $\downarrow$&Cog. $\downarrow$\\\midrule
  \rowcolor{lightgray} \multicolumn{9}{c}{\textbf{\textit{Referenced Results (Not Directly Comparable)}}} \\ \midrule
GPT-4V \cite{achiam2023gpt}$^\dagger$$^\sharp$ &3.49 & 0.28 &13.6 &7.3 &4.6 &67.1 &30.7 &2.6 \\ \midrule
LLaVA-v1.5-7B \cite{liu2024improved}$^\ddagger$$^\sharp$ & 2.11 & 0.54 & 53.6 & 25.2 & 7.8 & 51.0 & 36.4 & 4.2 \\
+ HACL \cite{jiang2024hallucination}$^\ddagger$ &2.13 &0.50 &- &- &- &- &- &- \\
+ POVID \cite{zhou2024aligning}$^\sharp$ &2.08 &0.56 &48.1 &24.4 &- &- &- &- \\
+ OPERA \cite{huang2024opera}$^\sharp$& 2.15 & 0.54 & 45.1 & 22.3  &- &- &- &- \\
+ VCD \cite{leng2024mitigating}$^\sharp$& 2.12 & 0.54 & 48.8 & 24.3 &- &- &- &- \\
+ EOS \cite{yue2024less}$^\ddagger$$^\sharp$ & 2.03 & 0.59 & 40.3 & 17.8 & 5.1 & 49.1 & 22.7 & 2.0 \\
+ HA-DPO \cite{zhao2023beyond}$^\ddagger$$^\sharp$ &1.97 &0.60 & 39.9 &19.9 &6.7 &49.8 &30.9 &3.3 \\
+ HALVA \cite{sarkar2024mitigating}$^\ddagger$ &2.25 &0.54 &- &- &6.6 &53.0 &32.2 &3.4 \\ \midrule
LLaVA-v1.5-13B \cite{liu2024improved}$^\dagger$ & 2.42 & - & 46.3 & 22.6 & 7.8 & 51.0 & 36.4 & 4.2  \\
+ RLHF-V \cite{yu2024rlhf}$^\dagger$ &2.81 &0.49 &12.2 &7.5 &6.3 &46.1 &25.1 &2.1 \\ 
+ HSA-DPO \cite{xiao2024detecting}$^\dagger$ & 2.61 & 0.48 & 5.2 & 3.2 & 2.1 & 47.3 & 13.4 & 1.2 \\
+ HALVA \cite{sarkar2024mitigating}$^\ddagger$ &2.58 &0.45 &- &- &6.4 &52.6 &30.4 &3.2 \\\midrule
Qwen-VL-Chat \cite{bai2023qwen}$^\dagger$ &2.89 &0.43 &36.0 &21.3 &6.6 &53.2 &31.0 &2.9 \\
+ Silkie-80K \cite{li2023silkie}$^\dagger$ &3.01 &0.41 &25.3 &13.9 &5.4 &55.8 &29.0 &2.0 \\ \midrule
 \rowcolor{lightgray} \multicolumn{9}{c}{\textbf{\textit{3B Multimodal LLMs}}} \\ \midrule
Bunny-v1.0-3B \cite{he2024bunny} &2.11 &0.58 &43.0 &8.9 &9.8 &\textbf{75.6} &64.9 &6.0 \\
+ DPO  &2.28 &0.56 &44.3 &7.6 &7.9 &74.1 &58.9 &4.8 \\
+ \MODEL  &\textbf{2.96} &\textbf{0.42} &\textbf{27.0} &\textbf{4.6} &\textbf{4.9} &67.4 &\textbf{37.7} &\textbf{2.4} \\ \midrule
 \rowcolor{lightgray} \multicolumn{9}{c}{\textbf{\textit{7B Multimodal LLMs}}} \\ \midrule
LLaVA-v1.5-7B \cite{liu2024improved}  &2.19 &0.57 &54.7 &15.9 &7.4 &51.8 &34.7 &4.1 \\
+ DPO  &2.14 &0.65 &49.0 &13.0 &6.5 &\textbf{55.1} &34.5 &\textbf{2.3} \\
+ \MODEL  &\textbf{2.39} &\textbf{0.54} &\textbf{35.7} &\textbf{9.8} &\textbf{4.4} &52.4 &\textbf{24.5} &2.4 \\
\bottomrule
\end{tabular}
\caption{\textbf{Main results} of Bunny-v1.0-3B and LLaVA-v1.5-7B trained with different preference optimization objectives. 
We report overall score and hallucination rate (HalRate) on MMHalBench, CHAIR scores at both response and object levels on Object HalBench, along with CHAIR scores, object coverage (cover.), hallucination rate (HalRate), and cognition (Cog.) on AMBER. The best result for each metric in each group is in bold.
For reference, we also provide additional results using various multimodal LLMs, preference data, and learning objectives, although these are not directly comparable. Results from contemporary work focusing on multimodal preference data: $^\dagger$\citet{xiao2024detecting}, $^\ddagger$\citet{sarkar2024mitigating}, and $^\sharp$\citet{yu2024rlaif}.}
\label{tab:main}
\end{table*}

\subsection{Anchored Preference Optimization}
\label{sec:anchored}

We also observe that the likelihood of the chosen response often decreases during the optimization process of DPO. This occurs because the standard DPO objective only encourages the model to learn a relative preference. Without further regularization, the model may reduce the likelihood of the chosen response to enlarge the likelihood gap between the chosen and rejected responses. This can harm model performance, as the chosen responses are often of high quality. To address this problem, we add an anchor to the preference optimization, forcing the reward of the chosen response to be higher than a specific value: \( \sigma(r(m_w, q, y_w) - \delta) \) with $\delta$ as the anchor value. The corresponding objective is 
\begin{equation*}
\small
\begin{split}
         \mathcal{L}_\text{AncPO} = -\log \sigma \Big(\beta \log \frac{\pi_{\theta}(y_w | m_w, q)}{\piref(y_w | m_w, q)} - \delta \Big).
\end{split}
\end{equation*}
In this way, we introduce absolute reward regularization to the preference optimization process, effectively avoiding the likelihood decrease of the chosen response.
The anchor is decided based on the data properties and expected model behavior. While we keep it simple in the default setting of \MODEL, one can always change the anchor values and set multiple anchors for different purposes.
The anchor can also be added to the rejected response in the opposite direction, forcing its reward to be lower than a specific value.
We compare other anchors in \Cref{sec:analysis}.

The objective of \MODEL is a combination of the standard DPO, conditional preference optimization, and anchored preference optimization:

\begin{small}
\begin{equation*}
\mathcal{L}_\text{\MODEL} = \mathcal{L}_{\text{DPO}_m} + \mathcal{L}_\text{CoPO} + \mathcal{L}_\text{AncPO}.
\end{equation*}
\end{small}

\section{Experiment}

In this section, we begin with the experimental setup (\Cref{sec:setup}). Then we present the main results on three benchmarks (\Cref{sec:result}) and the human evaluation (\Cref{sec:human}) of \MODEL. We further provide in-depth analysis (\Cref{sec:analysis}) and fine-grained results (\Cref{sec:fine-grained}). Finally, we conduct a qualitative study (\Cref{sec:qual}).

\subsection{Experimental Setup}
\label{sec:setup}

\stitle{Models}
We apply \MODEL on two multimodal LLMs in different sizes.
Bunny-v1.0-3B \cite{he2024bunny} is a 3B model building upon SigLIP  \cite{zhai2023sigmoid} and Phi-2 \cite{javaheripi2023phi}. It is pretrained on 2M image-text pairs and finetuned on 695K instruction tuning data. 
LLAVA-v1.5-7B \cite{liu2024improved} is a 7B model based on CLIP \cite{radford2021learning} and Vincuna \cite{chiang2023vicuna}. It is pretrained on 558K image-text pairs and finetuned on 665K instruction tuning data.

\stitle{Preference Data}
We sample 10K preference data from Silkie \cite{li2023silkie} with instructions from LLaVA-Instruct-150K \cite{liu2024improved} for training. The original Silkie dataset contains 80K preference data collected on 12 multimodal LLMs. While the original Silkie paper explores the effect of extreme data size, we follow the majority of prior works using around 10K data for preference optimization \cite{sun2023aligning,zhao2023beyond}. 

\stitle{Evaluation Benchmarks}
We evaluate the performance of \MODEL on three widely used benchmarks for multimodal LLMs with a special focus on hallucination. 
MMHalBench \cite{sun2023aligning} is a practical question answering benchmark containing eight question categories and 12 object topics. Following the official setting, we use GPT4 \cite{achiam2023gpt} to assess the overall quality of responses with a score betwen zero and six, and the hallucination rate.
Object HalBench \cite{rohrbach2018object} is a widely adopted benchmark to assess object hallucination. We follow the setting of \citet{yu2024rlhf} to augment the benchmark with eight diverse prompts and evaluating on 300 instances. We report the CHAIR scores \cite{rohrbach2018object} assessing hallucination rate of response level (CHAIR$_s$) and object level (CHAIR$_i$).
AMBER \cite{wang2023llm} is a multimodal LLM hallucination benchmark with fine-grained object annotation. We focus on the generative task consisting of 1K images. Using the official evaluation tool, we report a variant of CHAIR score, object coverage, rate of hallucinated responses, and hallucination rate overlapping with human cognition.

\begin{figure}[t]
    \centering
    \includegraphics[width=\linewidth]{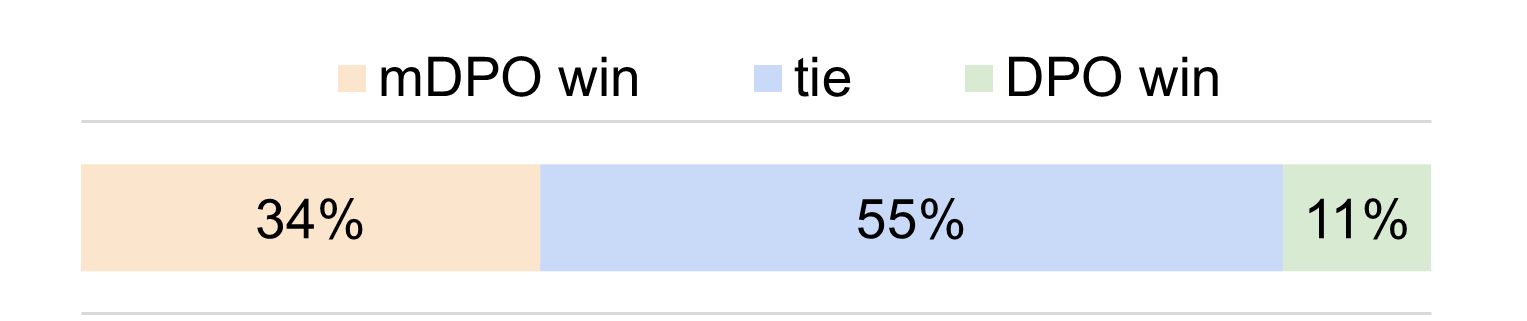}
    \caption{\textbf{Human evaluation} on MMHalBench.}
    \label{fig:human}
\end{figure}

\stitle{Baselines}
We primarily compare \MODEL with standard DPO. The standard DPO baseline shares the same training process, data, and hyper-parameters, despite different learning objectives.
We further provide the results of other multimodal LLMs for reference, although they are not directly comparable due to different base models, preference data, and alignment methods. This group contains GPT-4V \cite{achiam2023gpt}, LLaVA-v1.5-13B \cite{liu2024improved}, Qwen-VL-Chat \cite{bai2023qwen}, POVID \cite{zhou2024aligning}, HACL \cite{jiang2024hallucination}, OPERA \cite{huang2024opera}, VCD \cite{leng2024mitigating}, EOS \cite{yue2024less}, HA-DPO \cite{zhao2023beyond}, HALVA \cite{sarkar2024mitigating}, RLHF-V \cite{yu2024rlhf}, HSA-DPO \cite{xiao2024detecting}, and Silkie \cite{li2023silkie}. Some of them are contemporary works to ours.

\stitle{Implementation Details}
We train all the models for 3 epochs with a batch size of 32. We use a learning rate of 0.00001, a cosine learning rate scheduler, and a warmup ratio of 0.1. We set the $\beta$ of preference optimization to 0.1.
Following prior work \cite{zhao2023beyond,li2023silkie}, we use LoRA \cite{hu2021lora} to tune the model. Specifically, we set the $\alpha$ to 128 and rank to 64 for LoRA. \MODEL and standard DPO share the same configuration above. For \MODEL, we set $\delta=0$ by default.

\begin{figure}[t]
    \centering
    \includegraphics[width=0.9\linewidth]{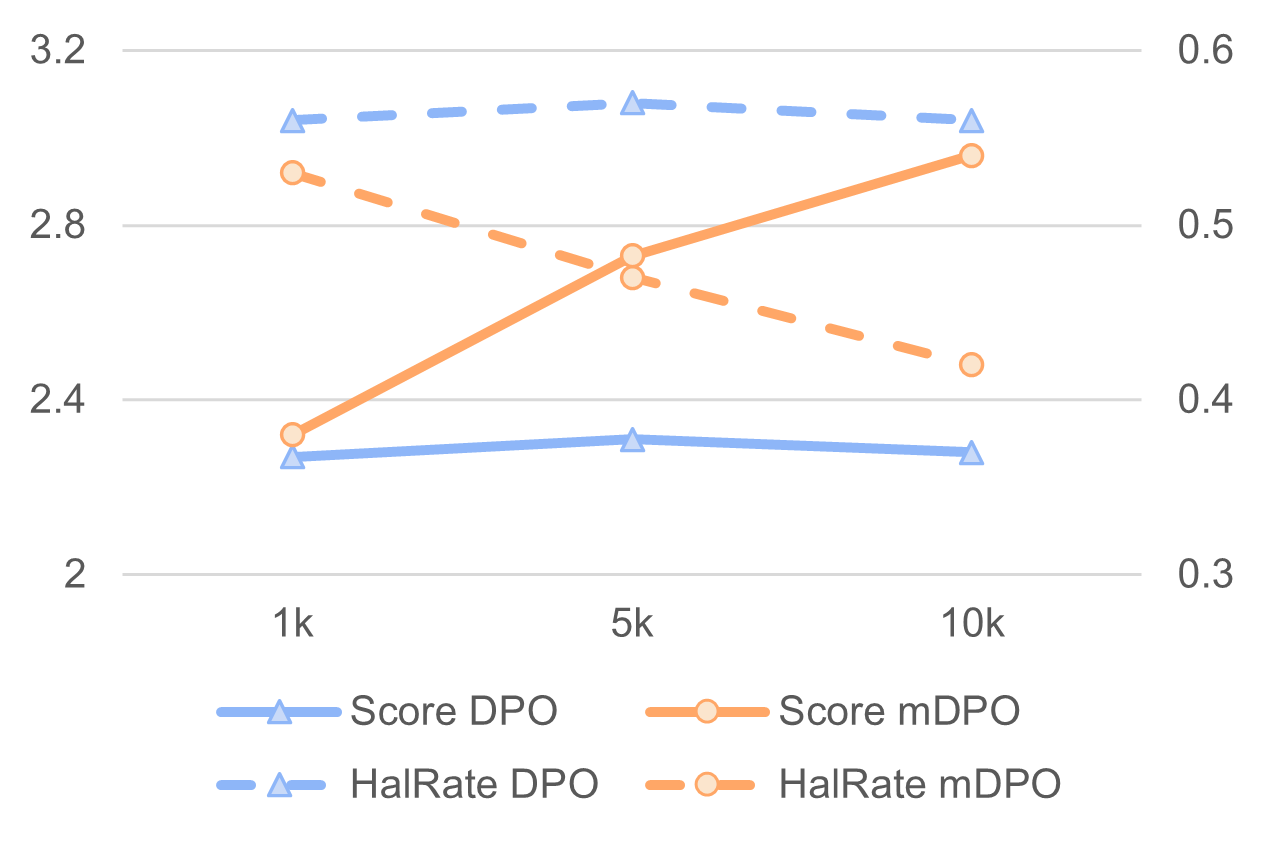}
    \caption{\textbf{Impact of data scale} on the performance of standard DPO and \MODEL, using Bunny as the base model. We assess the overall score and hallucination rate on MMHallBench. \MODEL is effective across different scales, whereas standard DPO does not exhibit a scaling effect in multimodal scenarios.}
    \label{fig:scale}
\end{figure}

\subsection{Main Results}
\label{sec:result}

\Cref{tab:main} presents the main results. On all three benchmarks, \MODEL consistently performs better than DPO for Bunny and LLaVA. Notably, \MODEL enhances the 3B model (Bunny) with 10K preference data to be comparable to a stronger 7B base model (Qwen-VL-Chat) trained with DPO on 80K data. The former preference data is only a subset of the latter. This result highlights that a proper objective can be more important than data scale and diversity in multimodal preference optimization. Moreover, \MODEL is specifically effective in reducing hallucination, which aligns with the objective of conditional preference optimization. While \MODEL may lead to a decrease in object coverage, this decrease is minor given the significant improvement in overall quality and reduction in hallucination.

\subsection{Human Evaluation}
\label{sec:human}
To further verify the effectiveness of \MODEL, we conduct human evaluation on MMHalBench, in which we ask domain experts to pick the better response generated by Bunny trained with either DPO or \MODEL. The results are presented in \Cref{fig:human}. Overall, responses from \MODEL are of better or same quality on 89\% instances compared to DPO. In contrast, DPO only achieves better performance on 11\% instances.

\begin{table}[t]\centering
\small
\setlength{\tabcolsep}{3.5pt}
\begin{tabular}{lcccc}\toprule
&\multicolumn{2}{c}{MMHalBench} &\multicolumn{2}{c}{Object HalBench} \\ \cmidrule(rl){2-3} \cmidrule(rl){4-5}
&Score &HalRate &CHAIR$_s$ &CHAIR$_i$ \\\midrule
\textbf{mDPO} &\textbf{2.96} &\textbf{0.42} &\textbf{27.0} &\textbf{4.6} \\
- conditional &2.36 &0.53 &40.3 &7.1 \\
- anchored  &2.50 &0.48 &34.3 &5.7 \\
- both (i.e., DPO) &2.28 &0.56 &44.3 &7.6 \\
\bottomrule
\end{tabular}
\caption{\textbf{Ablation results on \MODEL with conditional preference or/and anchored preference removed.} While both components are essential in \MODEL, anchored preference alone brings only slight improvement over DPO. This indicates that conditional preference is crucial in multimodal scenarios.}
\label{tab:ablation}
\end{table}

\subsection{Analysis}
\label{sec:analysis}

\begin{table}[t]\centering
\small
\setlength{\tabcolsep}{3.5pt}
\begin{tabular}{lcccc}\toprule
&\multicolumn{2}{c}{MMHalBench} &\multicolumn{2}{c}{Object HalBench} \\ \cmidrule(rl){2-3} \cmidrule(rl){4-5}
&Score &HalRate &CHAIR$_s$ &CHAIR$_i$ \\\midrule
Random image &2.81 &0.46 &40.7 &6.6 \\
\textbf{Crop 0-20\%} &\textbf{2.96} &\textbf{0.42} &\textbf{27.0} &\textbf{4.6} \\
Crop 20\%-50\% &2.92 &0.42 &33.7 &5.4 \\
MoCo v2 &2.82 &0.44 &32.3 &5.9 \\
\bottomrule
\end{tabular}
\caption{
\textbf{Comparison of strategies to create rejected images in \MODEL.} 
Among all the strategies, the default strategy in \MODEL, Cropping 0-20\% of the chosen images, retains some similarities with the original images but contains insufficient visual information, thereby providing effective preference optimization signals. In contrast, random images are too easy to identify, while MoCo v2's data augmentation \cite{chen2020improved} may not produce clearly worse images.
}
\label{tab:image}
\end{table}

\begin{table}[t]\centering
\small
\setlength{\tabcolsep}{3.5pt}
\begin{tabular}{lcccc}\toprule
&\multicolumn{2}{c}{MMHalBench} &\multicolumn{2}{c}{Object HalBench} \\ \cmidrule(rl){2-3} \cmidrule(rl){4-5}
Anchor&Score &HalRate &CHAIR$_s$ &CHAIR$_i$ \\\midrule
$\mathbf{y_w}$ &2.96 &0.42 &\textbf{27.0} &\textbf{4.6} \\
$y_w$ \& $y_l$  &\textbf{2.98} &\textbf{0.39} &29.3 &5.0 \\
$y_w$ \& $y_l $ \& $m_l$ &2.85 &0.4 &34.7 &6.1 \\
\bottomrule
\end{tabular}
\caption{
\textbf{Comparison of anchors used in \MODEL.}
\MODEL adds an anchor to regularize the $r(m_w, q, y_w)$ to be positive by default. Adding additional anchors to regularize the rewards of instances with rejected responses ($y_l$) or images ($m_l$) to be negative does not show an obvious improvement.
}
\label{tab:anchor}
\end{table}

\stitle{\MODEL is effective across different scales of preference data} We assess the overall score and hallucination rate on MMHallBench, as shown in \Cref{fig:scale}. We find that \MODEL is effective and consistently outperforms DPO across different data scales, demonstrating that our conditional preference method enhances multimodal preference optimization. Additionally, we observe that \MODEL's performance increases with the scale of data, while DPO does not exhibit a scaling effect. This indicates that \MODEL better utilizes multimodal preference data compared to DPO. Specifically, DPO struggles to fully leverage multimodal preference data, and its neglect of the visual modality cannot be mitigated by merely increasing the size of the preference data.

\stitle{Both designs in \MODEL are effective, with conditional preference being more crucial} We conduct an ablation study to evaluate the contributions of each component in \MODEL, as shown in \Cref{tab:ablation}. While both anchored preference and conditional preference enhance the overall performance of \MODEL, the results indicate that conditional preference leads to greater improvements than anchored preference. This suggests that conditional preference is the key factor in enhancing the effectiveness of DPO in multimodal scenarios, ensuring that the model better utilizes the visual modality.

\stitle{Using hard negative images for rejection improves preference optimization} We evaluate \MODEL with different methods for constructing the rejected image, as shown in \Cref{tab:image}. Among all the strategies, the default strategy in \MODEL, which involves cropping 0-20\% of the chosen images, consistently outperforms the others. This indicates that using hard negative images, which retain some similarities with the original images but also have parts erased, provides effective preference optimization signals. In contrast, random images are too easy to identify, while MoCo v2's data augmentation \cite{chen2020improved} is for creating similar images.

\stitle{Adding anchors to rejected responses or images brings litter improvement}
In \MODEL, we introduce an anchor to regularize $r(m_w, q, y_w)$ to be positive by default.
We also experimented with adding additional anchors to regularize $r(m_w, q, y_l)$ and $r(m_l, q, y_w)$ to be negative.
However, the additional anchors do not yield significant improvements.
The results, shown in~\Cref{tab:anchor}, indicate that only using the anchor on $r(m_w, q, y_w)$ is sufficient. Adding anchors to rejected responses or images may complicate the training process without providing clear advantages.

\begin{table*}[t]\centering
\small
\begin{tabular}{lccccccccc}\toprule
&overall &attribute &adversarial &comparison &counting &relation &environment &holistic &other \\\midrule
Bunny &2.11 &\textbf{3.92} &0.83 &\textbf{2.17} &2.33 &2.67 &2.25 &1.75 &1.00 \\
+ DPO &2.28 &3.25 &1.50 &1.42 &2.50 &2.67 &\textbf{4.25} &1.75 &0.92 \\
+ \MODEL &\textbf{2.96} &3.08 &\textbf{4.17} &2.00 &\textbf{3.50} &\textbf{3.25} &4.08 &\textbf{2.17} &\textbf{1.42} \\
\bottomrule
\end{tabular}
\caption{\textbf{Fine-grained results on MMHalBench} with a maximum score of six. \MODEL outperforms standard DPO on six out of eight types of questions, showing significant improvement particularly on adversarial questions with false premises about images.}
\label{tab:fine-grained}
\vspace{-0.5em}
\end{table*}

\subsection{Fine-grained Results}
\label{sec:fine-grained}

We further compare the  fine-grained results of DPO and \MODEL on MMHalBench. As shown in \Cref{tab:fine-grained}, among the eight question categories, \MODEL outperforms standard DPO on six of them. \MODEL shows significant improvement particularly on adversarial questions with false premises about images. \MODEL can identify the incorrect information in the question according to the image, while DPO fail to do so. These results also show the advantage of \MODEL under various practical scenarios. %
\subsection{Qualitative Study}
\label{sec:qual}

In \Cref{fig:qual}, we compare the 
When trained with standard DPO, Bunny often assumes the image description in the question is correct, responding accordingly, even if the question contains an adversarial premise regarding the image. In contrast, \MODEL identifies the false premise in the question by referencing the image. 
Moreover, Bunny trained with standard DPO may disregard the image and provide an educated guess for the answer. Conversely, \MODEL delivers a correct answer that is conditioned on the image.

\section{Related Work}

Reinforcement learning from human feedback~(RLHF; \citealt{christiano2017deep, ouyang2022training}) has proven to be an effective approach for aligning LLMs with human values. Direct preference optimization (DPO; \citealt{rafailov2024direct}), which involves directly optimizing LLMs based on human preferences, has been widely adopted in RLHF due to its strong performance and the elimination of the need for a separate reward model.
Significant efforts have been made to further enhance the efficacy and efficiency of DPO, which can be categorized into algorithmic and data-related advancements. On the algorithmic side, various approaches aim to improve the efficiency of DPO. For example, ORPO~\cite{hong2024reference} models preferences using an odds ratio and combines instruction fine-tuning and preference optimization into a unified training process. Methods such as CPO~\cite{xu2024contrastive}, TPO~\cite{saeidi2024triple}, and SimPO~\cite{meng2024simpo} simplify DPO by eliminating the use of a reference model, thereby reducing computational and memory overhead. 
Additionally, IPO~\cite{azar2024general} addresses the issue of reward overfitting in DPO.
On the data side, approaches such as KTO~\cite{ethayarajh2024kto} and NCA~\cite{chen2024noise} seek to overcome DPO’s requirement for paired preference data by designing optimization goals that can also utilize unpaired data. Iterative DPO~\cite{xu2023some, yuan2024self, xiong2024iterative} and SPPO~\cite{wu2024self} propose sampling preference data in an on-policy manner, achieving better results than off-policy DPO. 
WPO \cite{zhou2024wpo} adapts off-policy data to resemble on-policy data by reweighting preference pairs.
More closely related to our work, studies by \citet{park2024disentangling} and \citet{dong2024rlhf} address the reward hacking problem in textual preference optimization, where human preference may be biased towards longer outputs.
They propose to calibrate the rewards in DPO with respect to output length.
In this work, we discover a novel challenge in multimodal DPO, where preference optimization often neglects images. We then propose a solution to this problem through conditional preference optimization.

In multimodal scenarios, recent works mainly focus on creating multimodal preference data \cite{li2023silkie,zhao2023beyond,xiao2024detecting,zhou2024aligning,pi2024strengthening,sarkar2024mitigating,yu2024rlaif,deng2024enhancing}. These efforts include collecting human preference \cite{sun2023aligning,yu2024rlhf}, preference from advanced multimodal LLMs \cite{li2023silkie,yu2024rlaif}, and preference from the model to align itself \cite{deng2024enhancing}. 
In terms of learning objectives, recent works mainly follows DPO for LLMs \cite{li2023silkie,zhao2023beyond,zhou2024aligning}. Some also apply reinforcement learning \cite{sun2023aligning,jing2024fgaif} and contrastive learning \cite{sarkar2024mitigating,jiang2024hallucination}.
Our work studies an overlooked but crucial problem in the multimodal DPO objective.

\section{Conclusion}

We propose \MODEL, a preference optimization method dedicated to multimodal scenarios. \MODEL leverages conditional preference optimization to encourage multimodal LLMs to capture preference labels based on both visual and language cues. It further introduces anchored preference optimization to prevent the likelihood of preferred responses from decreasing. Experiments show that \MODEL consistently enhances multimodal LLM performance and reduces hallucination across different model sizes on three widely used benchmarks. 
Our method could be extended to reduce hallucinations and enhance trustworthiness across broader multimodal data and scenarios, including multiple images \citep{wang2024muirbench}, videos \citep{li2024llava}, in-context learning \citep{xu2024introspection}, and risk-sensitive domains \citep{chaves2024training}.

\section*{Acknowledgement}

We thank the anonymous reviewers for their valuable comments.
Fei Wang was supported by the Amazon ML Fellowship.
Muhao Chen was supported by the DARPA FoundSci Grant HR00112490370, the NSF of the United States Grant ITE 2333736, and an Amazon Research Award.

\section*{Limitation}
While we have conducted comprehensive experiments to show the effectiveness of \MODEL, there are still several limitations. First, experiments on more multimodal LLMs will provide further evidence on the advantages and disadvantages of \MODEL, specifically on models across various sizes and different architectures. Second, we focus on the unconditional preference problem in multimodal preference optimization. However, many contemporary studies have explored enhancing DPO from other perspectives, which may be complementary to ours. We leave the analysis of combining methods for future work. Third, while we have evaluated \MODEL on three benchmarks, they still represent a limited range of tasks and settings compared with the numerous scenarios in the real world. Further evaluation on more benchmarks can deepen our understanding of the proposed method.

\bibliography{reference}
\bibliographystyle{acl_natbib}

\end{document}